\documentclass[conference]{IEEEtran}
\usepackage[T1]{fontenc}
\usepackage[utf8]{inputenc}
\usepackage{amsmath,amssymb}
\usepackage{graphicx}
\usepackage{booktabs,tabularx,array}
\usepackage[hidelinks]{hyperref}
\hypersetup{pdftitle={When 10,000 Windows Are Not 10,000 Tests: Auditing Statistical Confidence in Sliding-Window Time-Series Classification},pdfauthor={Xinze Shi; Litian Zhang; Binrui Shi}}
\newcolumntype{Y}{>{\raggedright\arraybackslash}X}
\begin{document}
\title{When 10,000 Windows Are Not 10,000 Tests:\\Auditing Statistical Confidence in Sliding-Window\\Time-Series Classification}
\author{\IEEEauthorblockN{Xinze Shi, Litian Zhang$^{*}$, Binrui Shi}
\IEEEauthorblockA{Beijing University of Posts and Telecommunications\\
Beijing, China\\
\textsuperscript{*}Corresponding author:\\
Litian Zhang (\href{mailto:litianzhang@bupt.edu.cn}{litianzhang@bupt.edu.cn})}}
\maketitle

\AddToHookNext{shipout/foreground}{%
  \put(\dimexpr1in+\oddsidemargin\relax,-\dimexpr\paperheight-24pt\relax){%
    \makebox[0pt][l]{%
      \parbox[b]{\textwidth}{%
        \normalfont\fontsize{8}{9.5}\selectfont
        \noindent \textcopyright{} 2026 IEEE. Personal use of this material is permitted. Permission from IEEE must be obtained for all other uses, in any current or future media, including reprinting/republishing this material for advertising or promotional purposes, creating new collective works, for resale or redistribution to servers or lists, or reuse of any copyrighted component of this work in other works.
      }%
    }%
  }%
}

\begin{abstract}
Sliding-window classifiers are often evaluated on thousands of overlapping test windows, even though neighboring predictions share observations and remain nested within recordings and subjects. Subject-disjoint evaluation prevents one form of leakage but does not make those test windows independent. We present a practical audit that maps three claims---performance on observed recordings, future recordings from observed subjects, and unseen subjects---to explicit aggregation rules and established dependence-robust inference. At 75\% overlap, controlled simulations give 16.9\% Type-I error for IID observed-record inference and 7.2\% for session-centered Bartlett-HAC: a substantial improvement with residual miscalibration. Audits of frozen WISDM and HARTH predictions show that nearly fourfold growth in test rows yields only 1.75--1.94-fold variance-equivalent information growth. At that overlap, fixed-record paired Accuracy-difference intervals are 1.22--1.66 times the IID widths; this inflation is not universal at zero overlap. On HARTH, paired Accuracy-difference intervals include zero across three overlap settings, whereas Macro-F1 favors MiniROCKET. Independent recomputation, common-session checks, class-level results, and separately seeded calibration make the audit's scope and limitations inspectable. The resulting workflow distinguishes additional predictions from additional independent evidence.
\end{abstract}

\section{Introduction}
\begin{figure*}[t]
\centering
\includegraphics[width=.86\textwidth]{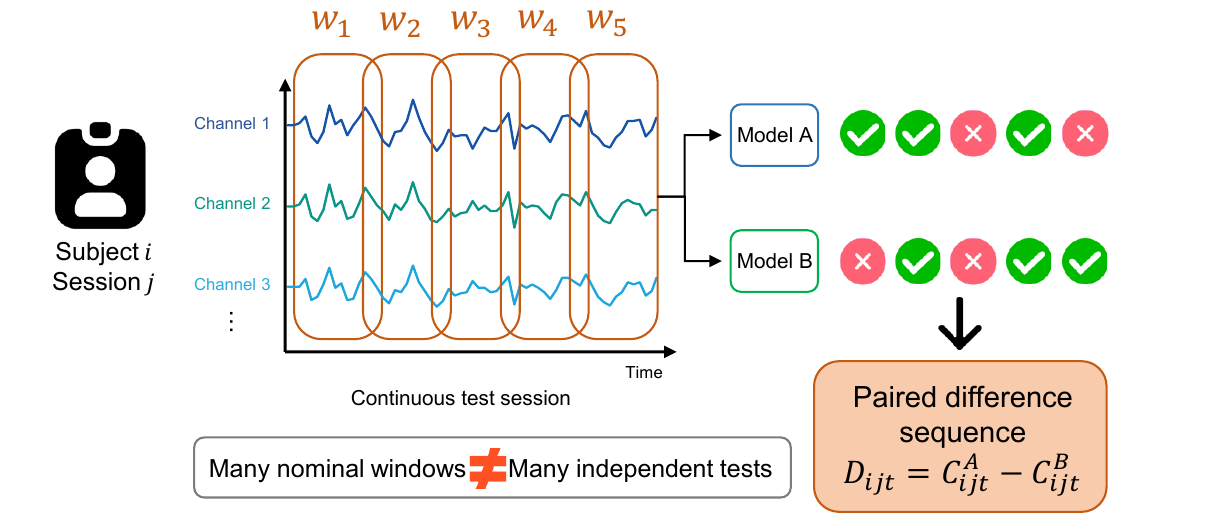}
\caption{Why nominal window count is not independent test count. Overlapping windows reuse the same multichannel samples; models are compared on matched windows, producing a temporally dependent paired contrast sequence.}
\label{fig:1}
\end{figure*}

Segmenting continuous recordings into fixed-length examples makes standard classifiers applicable to time-series data. Length and stride are usually selected for predictive or computational reasons \cite{r1}. At 75\% overlap, a recording produces approximately four times as many test rows as non-overlapping segmentation, but these are not four times as many independent tests. Adjacent windows reuse raw samples, errors persist, and predictions remain nested within recordings and subjects (Fig.~\ref{fig:1}). We examine this hierarchy on two wearable-sensor HAR datasets.

Subject-dependent splitting and overlapping windows can cause train--test leakage and inflated point performance \cite{r2}, \cite{r3}, \cite{r4}, \cite{r5}. We study test-comparison uncertainty after subject-disjoint evaluation: holding out subjects does not justify IID intervals from dependent test windows. Recent anomaly-detection work also shows that inference stride can alter point performance and rankings \cite{r6}; window generation is therefore part of the evaluation protocol.

The sampling claim also matters. ``Which model is better on these recorded windows?'', ``Which model is better on a future recording from the same people?'', and ``Which model is better for a new person?'' are different statistical questions. The estimands framework argues that evaluation should begin with a clearly defined target population and aggregation rule \cite{r7}. Recent work develops clustered inference for paired nonlinear classifier metrics \cite{r8} and intervals for nested classifier data \cite{r9}. These capabilities motivate our implementation. Our contribution is an executable mapping from a sliding-window claim to its aggregation rule, inference unit, eligibility checks, and reporting output (Table~\ref{tab:1}).

The audit separates temporal, session, and subject variation; checks calibration using the same formal estimator as the real-data analysis; and reports both Accuracy and Macro-F1 from frozen WISDM/HARTH predictions. Its contribution is the claim-to-analysis workflow and associated evidence checks.

\section{Related Work}
\textbf{Windowing and HAR validation.} Window size affects recognition latency and point performance \cite{r1}, while sample-generation and validation choices can strongly affect reported HAR accuracy \cite{r4}, \cite{r5}. Dehghani et al. isolated the interaction between subject splitting and overlapping windows \cite{r2}, \cite{r3}. These studies motivate subject-disjoint evaluation, but do not resolve the sampling uncertainty among dependent test windows once predictions have been generated.

\textbf{Uncertainty in model comparison.} Classical work warns that naive significance tests can be miscalibrated when errors share training or test data \cite{r10}, \cite{r11}, \cite{r12}. That literature primarily concerns variability across training/test splits. Our main estimand conditions on frozen out-of-fold predictions and asks how much evidence is contained in the dependent test sequence; a descriptive fold-level stability check is reported separately.

\textbf{Dependence-robust inference.} HAC covariance estimation \cite{r13}, moving-block and stationary bootstrap methods \cite{r14}, \cite{r15}, \cite{r16}, cluster-robust inference \cite{r17}, \cite{r18}, and hierarchical bootstrap \cite{r19} are established; we use them as components of a target-aligned audit whose contribution is the executable claim-to-analysis mapping, not a new generic variance estimator.

\begin{table}[t]
\caption{Capabilities demonstrated in the closest frameworks. ``Not shown'' describes the cited paper, not an impossibility.}
\label{tab:1}
\centering\footnotesize
\setlength{\tabcolsep}{3pt}
\begin{tabularx}{\columnwidth}{@{}>{\raggedright\arraybackslash}p{.19\columnwidth}Y@{}}
\toprule
Work & Demonstrated scope and relation to this audit\\
\midrule
Binette--Reiter \cite{r7} & Defines targets, aggregation, uncertainty, and reporting; a sliding-window three-target implementation is not shown.\\
Hong et al. \cite{r8} & Cluster sandwich, paired comparisons, nonlinear confusion-matrix metrics, and power; permits within-cluster dependence.\\
Anglin \cite{r9} & Nested-data metric intervals, F1 resampling, effective sample size and design degrees of freedom; sliding-window paired contrasts are not shown.\\
This audit & Maps temporal, session, and subject claims to weights and inference; checks paired windows, overlap eligibility, and relative information growth.\\
\bottomrule
\end{tabularx}
\end{table}

\section{Targets and Audit Protocol}
\subsection{Data hierarchy and paired contrasts}
Let subjects be indexed by $i$, sessions by $j$, and time-ordered windows by $t$. Classifiers $A$ and $B$ are evaluated on the same eligible windows. For model $m\in\{A,B\}$, let $\hat y^m_{ijt}$ be its predicted label and $y_{ijt}$ the ground-truth label on window $(i,j,t)$, and define the window-level correctness indicator
\[
C^m_{ijt}=\mathbf{1}\{\hat y^m_{ijt}=y_{ijt}\},
\]
For Accuracy, the paired contrast is the matched difference of correctness indicators,
\begin{equation}
\begin{aligned}
D_{ijt}&=C^A_{ijt}-C^B_{ijt}\\
&=\mathbf{1}\{\hat y^A_{ijt}=y_{ijt}\}-\mathbf{1}\{\hat y^B_{ijt}=y_{ijt}\}.
\end{aligned}
\label{eq:1}
\end{equation}
Positive values favor $A$. Pairing is essential: separately estimating two metric variances discards their covariance on the shared test windows. The audit requires matching subject, session, time, label, fold, and window-boundary keys for both models before inference.

\subsection{Three evaluation targets}
\begin{figure*}[t]
\centering
\includegraphics[width=\textwidth]{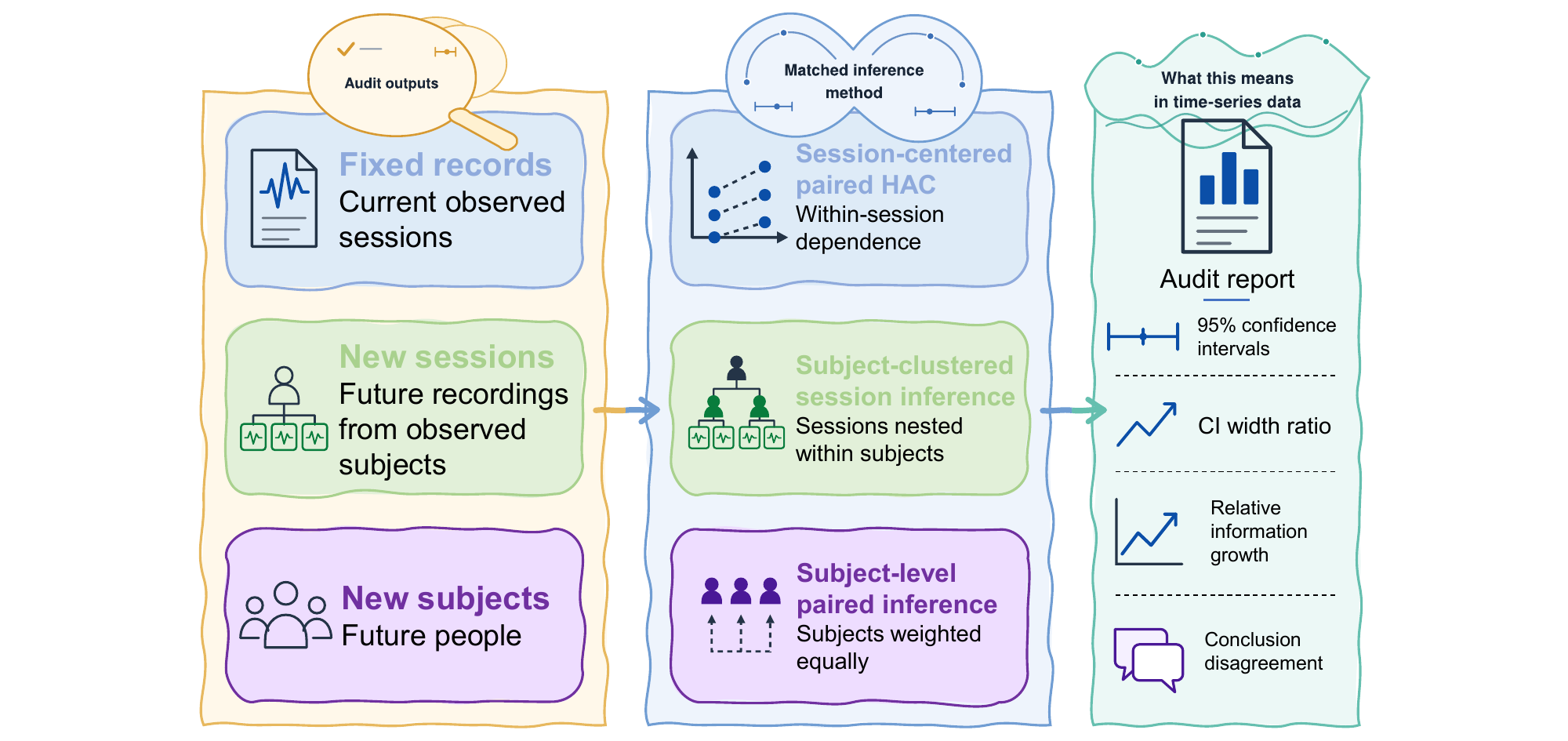}
\caption{Estimand-aware audit workflow. The intended generalization target determines the highest independent unit and therefore the matched inference method. The report exposes confidence intervals, CI width ratios, relative information growth, and conclusion disagreement.}
\label{fig:2}
\end{figure*}

We use ``target'' in the estimand sense of a population and aggregation rule \cite{r7}. Figure~\ref{fig:2} summarizes the decision path.

\textbf{Observed-record target.} Sessions and subjects are fixed and all observed windows are pooled:
\begin{equation}
\Delta_{\mathrm{obs}}=\frac{1}{N}\sum_{i,j,t}D_{ijt}.
\label{eq:2}
\end{equation}
Subject and session identities are fixed. Uncertainty concerns a finite temporally dependent realization within those sessions, not new recordings or people.

\textbf{New-session target.} Let $R$ count contributing sessions and $\bar D_{ij}$ be a session mean. Sessions are equally weighted, while sessions from the same subject remain dependent:
\begin{equation}
\Delta_{\mathrm{sess}}=\frac{1}{R}\sum_{i,j}\bar D_{ij}.
\label{eq:3}
\end{equation}
We use a subject-clustered sandwich calculation on session means; for nonlinear metrics, each session confusion matrix is divided by its window count before aggregation and paired subject-cluster resampling.

\textbf{New-subject target.} Let $\bar D_i$ be the subject-level metric difference. Subjects receive equal weight:
\begin{equation}
\Delta_{\mathrm{subj}}=\frac{1}{I}\sum_i\bar D_i.
\label{eq:4}
\end{equation}
We use paired inference across subjects. A subject with more windows does not receive more weight merely because its recording is longer.

\subsection{Observed-record HAC inference}
For observed-record Accuracy, paired differences are centered within each session, $u_{ijt}=D_{ijt}-\bar D_{ij}$, so between-session difficulty is not misinterpreted as within-session temporal dependence. Because the observed sessions and their process-level mean differences are conditioned on for this target, between-session mean heterogeneity is not part of the conditional temporal sampling variance. With Bartlett weights $w_h=1-h/(K+1)$ and the finite-sample factor $c=N/(N-R)$ where $R$ is the total number of contributing sessions (one estimated mean per session),
\begin{equation}
\widehat{\operatorname{Var}}(\Delta_{\mathrm{obs}})=\frac{c}{N^2}\sum_{i,j}\left[\sum_t u^2_{ijt}+2\sum_{h=1}^{K}w_h\sum_t u_{ijt}u_{ij,t+h}\right],
\label{eq:5}
\end{equation}
where $N$ is the pooled number of windows across the $R$ sessions. Lag products never cross sessions. On the resampled grid, lag $h$ pairs retained starts separated by exactly $hS$; HARTH purity gaps remain gaps. Autocorrelation and bandwidth diagnostics use this same lag definition. Mechanical overlap supplies a minimum correlation range
\begin{equation}
K_0=\left\lceil\frac{L}{S}\right\rceil-1,
\label{eq:6}
\end{equation}
where $L$ is window length and $S$ is stride. A deterministic length-based Newey--West count $K_{\mathrm{NW}}=\lfloor4(\bar T/100)^{2/9}\rfloor$ is computed from the mean session length $\bar T$. The frozen default, informed by earlier synthetic calibration, is $K_{\mathrm{main}}=\max(K_{\mathrm{NW}},2K_0)$; $K_0$ and $4K_0$ are reported as sensitivity settings. The implementation caps the global bandwidth at the largest available within-session lag; shorter sessions contribute only legal pairs. Equation~(\ref{eq:5}) is the classical Newey--West/Bartlett construction \cite{r13}, specialized to paired sequences separated by session.

Macro-F1 is a nonlinear functional of the complete confusion matrix, not a mean of per-window ``F1 scores.'' For observed records we use a confusion-matrix influence-function HAC, with a session-aware block bootstrap that recomputes Macro-F1 in each replicate as a sensitivity check. New-session Macro-F1 is computed from the mean of session-normalized confusion matrices, not the mean of session F1 values; each subject resample retains its sessions and recomputes this mean. New-subject inference averages subject-specific Macro-F1 differences. All paths use the fixed dataset class set (18 WISDM, 12 HARTH), assigning zero F1 to a zero denominator, including classes absent within a subject. The influence-function construction follows smooth-functional metric inference \cite{r8}.

Fixed-record HAC intervals use $z_{.975}=1.960$, with no finite $t$ degrees of freedom. New-session Accuracy uses variance $[I/(I-1)]\sum_i[\sum_j(\bar D_{ij}-\Delta_{\mathrm{sess}})]^2/R^2$ and $t_{I-1}$; equal-subject intervals use $s_d/\sqrt{I}$ and $t_{I-1}$. Hence HARTH uses 22 subject clusters, df=21 and critical value 2.080; WISDM uses 51, df=50 and 2.009. New-session Macro-F1 uses 1,999 paired subject-cluster percentile replicates, without a $t$ reference. These finite-sample choices do not resolve all small-cluster problems \cite[Sec.~VI]{r18}.

\subsection{Audit outputs and interpretation}
For overlap level $r$ relative to a non-overlapping baseline, the audit reports
\begin{equation}
G_{\mathrm{info}}(r)=\frac{\widehat{\operatorname{Var}}_{\mathrm{dep}}(0)}{\widehat{\operatorname{Var}}_{\mathrm{dep}}(r)}.
\label{eq:7}
\end{equation}
This is variance-equivalent \emph{relative} information growth, not a universal absolute effective sample size. The target-aligned/IID width ratio is $W_{\mathrm{dep}}/W_{\mathrm{IID}}$ at a fixed overlap. The report also includes global and session-centered autocorrelation, bandwidth sensitivity, and a conclusion-disagreement flag when IID and target-aligned intervals support different decisions.

\subsection{From a scientific claim to an executable audit}
Declare target, metric, and difference direction before inference. Input checks reject mismatched or duplicate window keys, missing predictions, non-monotone original order, invalid boundaries, and subjects assigned to multiple test folds. Macro-F1 retains fixed class indexing in every resample. Information growth must disclose baseline eligibility: the original HARTH calculation used each overlap's contributing sessions; the common-session check below is additional.

\begin{table}[!ht]
\caption{Minimum fields in a dependence-aware audit report.}
\label{tab:2}
\centering\footnotesize
\setlength{\tabcolsep}{3pt}
\begin{tabularx}{\columnwidth}{@{}>{\raggedright\arraybackslash}p{.24\columnwidth}Y@{}}
\toprule
Field & Required interpretation\\
\midrule
Target & Fixed records, future sessions, or future subjects\\
Metric & Accuracy, Macro-F1, and the aggregation rule\\
Hierarchy & Subject/session identifiers and the highest independent unit\\
Window geometry & Length, stride, overlap, purity, and boundary policy\\
Estimate & Paired point difference on identical eligible windows\\
Uncertainty & Method, bandwidth/block rule, confidence level, and unit count\\
Diagnostics & IID/dependent widths, width ratio, information growth, and sensitivity\\
\bottomrule
\end{tabularx}
\end{table}

The report retains IID inference as a diagnostic reference. Disagreement between intervals identifies a target or uncertainty issue, rather than selecting a winner automatically.

\section{Experimental Design}
\subsection{Controlled calibration}
We generate paired binary correctness sequences with known marginal accuracies and controlled dependence at subject, session, raw-sample, and model-pair levels. Each subject has two sessions of fixed raw duration. Window length is $L=64$ raw samples with 0, 50, or 75\% overlap. Null experiments set both marginal accuracies to 0.8. We use 1,000--2,000 Monte Carlo test sets per condition and report empirical Type-I error with Monte Carlo intervals, following simulation-study guidance \cite{r20}. Conditions include overlap-only dependence, extended AR(1) dependence, subject/session random effects, nonzero differences, subject-count sensitivity, and paired-model correlation. The original multistage hierarchical bootstrap is retained as a diagnostic because pilot calibration found it over-conservative for the fixed-record target.

\subsection{Real data, models, and frozen comparisons}
WISDM contains smartphone and smartwatch streams from 51 subjects performing 18 activities at 20 Hz \cite{r21}; we use the phone accelerometer. HARTH contains professionally annotated free-living recordings from 22 subjects wearing thigh and lower-back accelerometers at 50 Hz \cite{r22}; we use the thigh sensor. Session boundaries are dataset-specific and fixed before analysis. For WISDM, a session is a maximal single-activity run within a subject's phone-accelerometer recording, split at activity changes, non-monotonic timestamps, or gaps exceeding five times the subject's median sampling interval; windows are therefore single-activity (label purity 1.0). For HARTH, a session is a maximal run of valid rows within a subject recording, split at excluded invalid rows, non-monotonic timestamps, or gaps exceeding one second, and a 10-s window is retained only if its majority label covers at least 0.8 of the window. Windows never cross a session boundary in either dataset.

Table~\ref{tab:3} summarizes the final window settings and the number of sessions contributing predictions at each tested overlap. The frozen boundary policy yields 934 WISDM and 591 HARTH segments; after window-length and purity checks, 933 WISDM sessions contribute at all tested overlaps, whereas HARTH contributes 393, 403, and 409 sessions at 0\%, 50\%, and 75\% overlap, respectively.

\begin{table}[!ht]
\caption{Real-data configurations after the frozen boundary and eligibility checks.}
\label{tab:3}
\centering\fontsize{8.5}{10.8}\selectfont
\setlength{\tabcolsep}{3pt}
\begin{tabular*}{\columnwidth}{@{\extracolsep{\fill}}lrrrrr@{}}
\toprule
Dataset & Window & Subj. & Rate & Test overlap & Used sessions\\
\midrule
WISDM & 5 s & 51 & 20 Hz & 0/75\% & 933/933\\
WISDM & 10 s & 51 & 20 Hz & 0/50/75\% & 933/933/933\\
HARTH & 10 s & 22 & 50 Hz & 0/50/75\% & 393/403/409\\
\bottomrule
\end{tabular*}
\end{table}

We compare a Random Forest using statistical and spectral features with MiniROCKET followed by validation-selected ridge classification \cite{r23}. Subject-level five-fold out-of-fold (OOF) prediction ensures that a subject occurs in exactly one test fold, with validation subjects drawn only from the corresponding training fold. Training windows use 50\% overlap. Test windows use 0, 50, and 75\% overlap. WISDM uses 10-s windows plus a prespecified and frozen 5-s sensitivity analysis; HARTH uses 10-s windows. All overlap settings use the same raw test sessions and held-out people. Retained windows and contributing sessions can nevertheless change after window-wise purity filtering. WISDM new-session claims concern new single-activity bouts under its segmentation rule; HARTH claims concern new continuous multi-activity stretches. They are different recording populations.

All analyses start from frozen prediction files. The artifact checks key uniqueness, matched rows, monotone time within session, fold/subject isolation, and window geometry before running inference. Random seeds, folds, feature settings, MiniROCKET parameters, and ridge search ranges are recorded in machine-readable manifests. The original real-data settings remain frozen; the additional analyses described below were conducted after review with separately locked configurations.

\begin{figure*}[t]
\centering
\includegraphics[width=.95\textwidth]{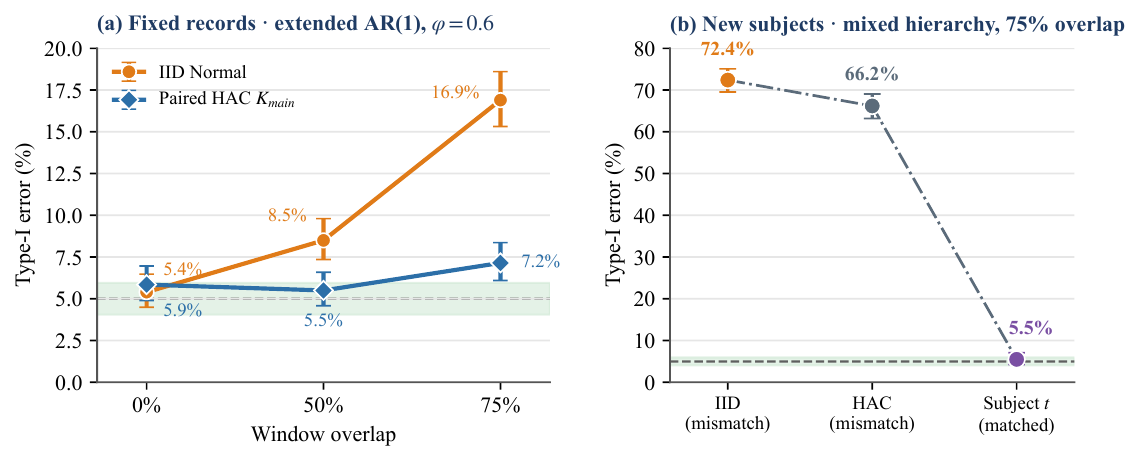}
\caption{Monte Carlo null calibration with 95\% Monte Carlo intervals. For the fixed-record target, the formal session-centered Bartlett-HAC procedure remains close to nominal at 0--50\% overlap and substantially reduces, but does not eliminate, anti-conservatism at 75\% overlap. For the new-subject target, window-level IID and within-session HAC inference remain severely target-misaligned, whereas equal-subject paired inference uses the matched sampling unit.}
\label{fig:3}
\end{figure*}

\subsection{Frozen comparisons and sensitivity grid}
Across overlaps, folds, training and validation subjects, fitted parameters, raw test recordings, boundaries, and purity rules remain fixed. The frozen 5-s WISDM analysis changes window geometry; all analyses use the archived predictions. The frozen sensitivity grid spans HAC bandwidth, Macro-F1 influence-function versus session-aware paired block bootstrap, subject count, session heterogeneity, and cross-model correlation. Five OOF folds provide only a descriptive training-set stability check, not five independent inference units.

The final historical HAC recalculation reused the earlier simulation seeds. Existing records do not establish independence between bandwidth selection and reported calibration. We therefore additionally locked the rules before generating new, independently seeded test sets; paired methods share evaluation samples. These post-review null and +2-percentage-point alternatives report Type-I error, coverage, power, width, and Monte Carlo standard errors (MCSE).

\section{Results}
\subsection{Target alignment determines calibration}
Figure~\ref{fig:3}(a) shows that IID Type-I error rises from 5.4\% at 0\% overlap to 8.5\% at 50\% and 16.9\% at 75\%. The formal session-centered Bartlett-HAC procedure yields 5.9\%, 5.5\%, and 7.2\%, respectively: it remains close to nominal at 0--50\% overlap and substantially reduces, but does not eliminate, anti-conservatism at the densest overlap. For a new-subject target under subject/session heterogeneity (Fig.~\ref{fig:3}(b)), IID window inference reaches 72.4\% Type-I error and the formal within-session HAC remains severely misaligned at 66.2\%, whereas equal-subject paired inference is 5.5\%. Thus, ``accounting for dependence'' is insufficient unless the correction is aligned with the intended highest independent unit.

\subsection{Overlap multiplies rows faster than information}
\begin{figure}[!ht]
\centering
\includegraphics[width=\columnwidth]{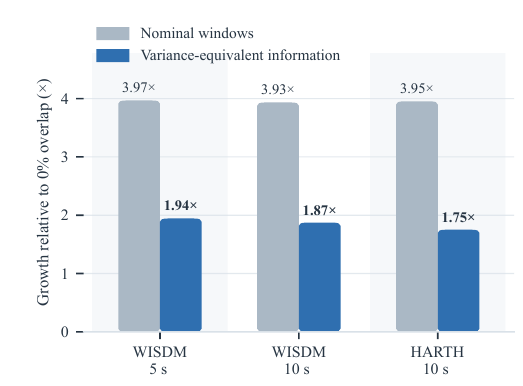}
\caption{Nominal test-window growth versus variance-equivalent information growth from 0\% to 75\% overlap. Information growth is the inverse session-centered Bartlett-HAC variance ratio relative to the 0\% setting; HARTH here uses each setting's eligible sessions.}
\label{fig:4}
\end{figure}

At 75\% overlap, nominal test-window counts grow by a factor of 3.93--3.97 across WISDM 5 s, WISDM 10 s, and HARTH 10 s. Dependence-aware variance decreases much less: variance-equivalent information grows by factors of 1.94, 1.87, and 1.75, respectively (Fig.~\ref{fig:4} and Table~\ref{tab:4}). These ratios describe evaluation information, not the predictive value of overlap.

\begin{figure}[t]
\centering
\includegraphics[width=\columnwidth]{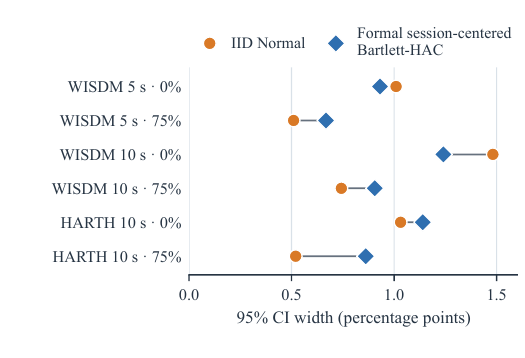}
\caption{Absolute 95\% confidence-interval widths for fixed-record paired Accuracy differences under IID-Normal and formal session-centered Bartlett-HAC inference. Table~\ref{tab:4} reports HAC/IID ratios.}
\label{fig:5}
\end{figure}

At dense overlap, target-aligned paired Accuracy-difference intervals are 1.22--1.66$\times$ wider than IID intervals (absolute widths in Fig.~\ref{fig:5}; ratios in Table~\ref{tab:4}). The magnitude depends on dataset, window length, label persistence, model errors, and session composition; a universal ``divide by four'' correction is not justified.

\begin{table}[t]
\caption{Fixed-record paired Accuracy-difference uncertainty at 75\% overlap. Widths are percentage points.}
\label{tab:4}
\centering\footnotesize
\begin{tabular*}{\columnwidth}{@{\extracolsep{\fill}}lccc@{}}
\toprule
Setting & Nom./Info. & IID/HAC width & HAC/IID\\
\midrule
WISDM 5 s & 3.97/1.94$\times$ & .510/.668 & 1.31$\times$\\
WISDM 10 s & 3.93/1.87$\times$ & .743/.906 & 1.22$\times$\\
HARTH 10 s & 3.95/1.75$\times$ & .520/.862 & 1.66$\times$\\
\bottomrule
\end{tabular*}
\end{table}

The sub-unity 0\%-overlap ratios decompose cleanly. Writing $W_{\mathrm{SC0}}$ for the session-centered lag-0 interval (same centering and finite-sample correction, no lag terms), $W_{\mathrm{HAC}}/W_{\mathrm{IID}}=(W_{\mathrm{SC0}}/W_{\mathrm{IID}})\times(W_{\mathrm{HAC}}/W_{\mathrm{SC0}})$, a centering ratio times a dependence ratio: $0.83\times1.11=0.92$ (WISDM 5 s) and $0.81\times1.03=0.84$ (WISDM 10 s). Session-centering narrows the interval by removing between-session heterogeneity, while the dependence factor exceeds one because adjacent non-overlapping windows remain temporally dependent.

\subsection{Metric Choice Changes the Scientific Conclusion}
\begin{figure*}[t]
\centering
\includegraphics[width=\textwidth]{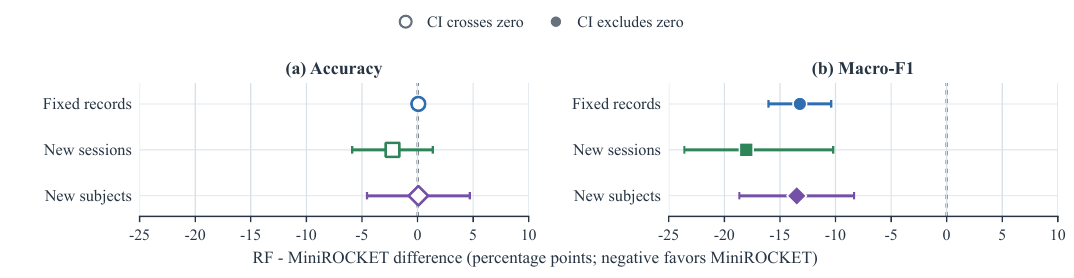}
\caption{HARTH-only RF$-$MiniROCKET differences with 95\% intervals (10-s windows, 75\% overlap) from the same frozen predictions. Accuracy and Macro-F1 support different conclusions; negative values favor MiniROCKET. Table~\ref{tab:6} reports complete cross-dataset results.}
\label{fig:6}
\end{figure*}

At 75\% overlap, HARTH paired Accuracy-difference intervals include zero for all three targets, while Macro-F1 favors MiniROCKET by 13.2--18.0 points. WISDM favors MiniROCKET for both metrics and all targets, including the frozen 5-s sensitivity. Figure~\ref{fig:6} shows HARTH; Table~\ref{tab:6} gives cross-dataset results. Metric weighting and target weighting are separate choices.

Fixed-record Accuracy weights every eligible window, so longer sessions, common activities, and window-rich subjects contribute more; Macro-F1 instead gives each class equal weight, emphasizing rare activities. Target weighting is separate---new-session inference equalizes sessions, new-subject inference equalizes people---so metric weighting and target weighting are distinct dimensions.

Aggregation must also be applied inside resampling: the new-session analysis averages session-normalized confusion matrices within each subject-cluster resample, while new-subject inference averages subject-specific metric differences. Accuracy reduces to weighted means, but nonlinear metrics such as Macro-F1 do not.

Metrics must be selected from the application objective before inspection: Accuracy reflects frequency-weighted cost, while Macro-F1 reflects a class-balanced objective. Significance is not a metric-selection rule.

Independent recomputation confirms Table~\ref{tab:6}. At 75\% overlap, HARTH's pooled and equal-subject Accuracy differences are respectively +0.073918641805 and +0.070878431669 percentage points; both round to +0.07. Unequal subject window counts do not force distinct displayed means.

\begin{table}[t]
\caption{HARTH overlap audit. Purity summaries weight retained windows equally. Difference point estimates are percentage points; complete CIs accompany the audit.}
\label{tab:5}
\centering\footnotesize
\begin{tabular*}{\columnwidth}{@{\extracolsep{\fill}}lrrr@{}}
\toprule
Overlap & 0\% & 50\% & 75\%\\
\midrule
Candidate windows & 12,250 & 24,271 & 48,296\\
Retained windows & 10,610 & 21,056 & 41,938\\
Retained (\%) & 86.61 & 86.75 & 86.84\\
Sessions / subjects & 393/22 & 403/22 & 409/22\\
Mean purity & .990389 & .990589 & .990705\\
Fully pure (\%) & 91.31 & 91.63 & 91.69\\
\bottomrule
\end{tabular*}
\vspace{1pt}
\begin{tabular*}{\columnwidth}{@{\extracolsep{\fill}}llrr@{}}
\toprule
Overlap & \multicolumn{1}{l}{Target} & Accuracy & Macro-F1\\
\midrule
0\% & \multicolumn{1}{l}{Fixed} & +0.17 & -11.87\\
0\% & \multicolumn{1}{l}{Session} & -1.38 & -16.08\\
0\% & \multicolumn{1}{l}{Subject} & +0.22 & -11.42\\
50\% & \multicolumn{1}{l}{Fixed} & +0.17 & -13.21\\
50\% & \multicolumn{1}{l}{Session} & -1.86 & -18.63\\
50\% & \multicolumn{1}{l}{Subject} & +0.20 & -12.88\\
75\% & \multicolumn{1}{l}{Fixed} & +0.07 & -13.21\\
75\% & \multicolumn{1}{l}{Session} & -2.25 & -18.03\\
75\% & \multicolumn{1}{l}{Subject} & +0.07 & -13.49\\
\bottomrule
\end{tabular*}
\end{table}

\begin{table*}[t]
\caption{Cross-dataset RF$-$MiniROCKET differences in percentage points with 95\% confidence intervals (10-s windows, 75\% overlap). Negative values favor MiniROCKET. Each column uses target-matched inference.}
\label{tab:6}
\centering\fontsize{9.5}{11.4}\selectfont
\begin{tabular*}{\textwidth}{@{\extracolsep{\fill}}llrrr@{}}
\toprule
Dataset & Metric & Fixed records & New sessions & New subjects\\
\midrule
WISDM & Accuracy & -9.65 [-10.10, -9.20] & -9.75 [-12.31, -7.18] & -9.38 [-11.81, -6.95]\\
WISDM & Macro-F1 & -8.39 [-8.85, -7.93] & -8.53 [-11.05, -6.08] & -8.81 [-11.08, -6.53]\\
HARTH & Accuracy & +0.07 [-0.36, +0.50] & -2.25 [-5.87, +1.37] & +0.07 [-4.55, +4.69]\\
HARTH & Macro-F1 & -13.21 [-16.03, -10.39] & -18.03 [-23.60, -10.21] & -13.49 [-18.65, -8.33]\\
\bottomrule
\end{tabular*}
\end{table*}

\textbf{HARTH overlap analysis.} An additional analysis after review examined the frozen eligibility rules. Windows contain 500 samples, start at session index zero, and use strides 500/250/125 with purity at least 0.8. The three grids are nested: all 31,328 retained windows added from 0\% to 75\% start at positions never generated on the coarse grid; they were not rejected coarse-grid windows. Their mean purity (0.990812) is slightly higher than that of shared windows (0.990389), and these summaries do not support a claim that the added positions are systematically less pure or harder.

Median purity is 1 at every overlap (5th percentiles: .896/.900/.902). Every paired Accuracy-difference CI includes zero and every Macro-F1-difference CI favors MiniROCKET, although point differences vary with overlap (Table~\ref{tab:5}).

The original information-growth calculation used 393/403/409 contributing HARTH sessions. An additional check recomputes all weights and variances on the 393 common sessions. At 75\%, $G_{\mathrm{info}}$ becomes 1.750128 versus 1.751439 originally; both round to 1.75. However, the equal-session paired Accuracy difference moves from $-2.25$ [$-5.87$, $+1.37$] to $-1.49$ [$-4.51$, $+1.52$] points. Thus, session weights matter despite similar information growth. This check fixes the recording set without intersecting windows or controlling all purity, class, or position differences.

At 75\% overlap, sitting supplies 52.04\% of HARTH windows. RF has 672 more correct sitting predictions and 415 more for seated cycling, partly offsetting MiniROCKET's gains elsewhere. Conversely, stair ascent/descent account for only 1.08\%/0.59\% of windows, but their F1 scores are .348/.186 for RF and .662/.752 for MiniROCKET. These two classes contribute 7.34 percentage points of MiniROCKET's 13.21-point pooled Macro-F1 advantage. This is a class-specific tradeoff, not evidence that every minority class benefits equally; full precision, recall, F1, support, and confusion matrices are supplied.

\subsection{Robustness and additional calibration}
The historical WISDM 5-s length-stress test gives HAC Type-I error 8.9\% versus IID 18.2\% at 75\% overlap, with HAC coverage 91.2\%. This identifies residual sensitivity to recording lengths. The single-file CLI implements Accuracy inference for all three targets; the independent audit script additionally supports Macro-F1 and information growth across overlaps. Frozen predictions, folds, configurations, and input tests define the reproducible analysis.

For WISDM 5 s at 75\% overlap, the influence-function Macro-F1 interval and a 1,999-replicate session-aware block bootstrap are nearly identical. Bandwidth sensitivity changes widths but not the reported real-data model conclusions.

\textbf{Independent calibration and power.} As an additional analysis after review, we generated 1,000 independent test sets per condition under separately saved seeds and fixed rules (24 conditions including null and +2-point alternatives). For the Fig.~\ref{fig:3}(a) designs, new IID rejection rates are 5.2/8.7/16.9\% and formal HAC rates are 5.5/6.4/7.9\% at 0/50/75\% overlap. In the hierarchical design, IID/HAC/subject-$t$ reject 73.2/66.3/4.6\%. MCSE is $\sqrt{\hat p(1-\hat p)/M}$ with the condition's actual $M$; the historical 7.15\% HAC result has MCSE 0.58 percentage points ($M=2000$). Paired method differences use replicate-wise covariance.

With a true +2-point difference, formal HAC power in the balanced AR(1) designs is 67.7/85.6/93.9\%; corresponding coverage is 93.1/94.6/92.8\%. Hierarchical equal-subject power is 14.1\% with 94.9\% coverage. The higher rejection of target-misaligned methods cannot be read as a power advantage because their null rejection is excessive.

The added per-session sensitivity uses
\[
K_{ij}=\min\left\{T_{ij}-1,\max\left[\left\lfloor4(T_{ij}/100)^{2/9}\right\rfloor,2K_0\right]\right\}.
\]
It retains short sessions and the same centering and finite-sample correction; available lags cap the mechanical-overlap requirement. Both rules are evaluated on identical draws. At 75\%, the $2K_0=6$ floor dominates: neither rule changes rejection counts in the new WISDM 5-s/HARTH 10-s length-template stress tests (6.7/6.6\%, respectively), although widths differ slightly. At lower overlap the complete sensitivity grid reports the actual changes. This is not evidence that per-session bandwidth cures the residual miscalibration, and the original default is retained.

The new WISDM length-stress result (6.7\%) differs from the historical 8.85\% by 2.12 combined MCSEs. The two rates are retained as separate Monte Carlo outcomes. Original rule-selection/calibration independence remains unverified. The new length-template simulations reproduce lengths, not HARTH purity gaps or class transitions.

\section{Discussion and Reporting Guidance}
A $p$-value or CI needs an explicit target population, metric aggregation, highest independent unit, and dependence structure. Figure~\ref{fig:2} links these declarations to the audit.

Without subject identifiers, only fixed-record and recording-level analyses remain. One recording per subject does not identify a same-subject new-session target; a single series supports only a conditional temporal target without additional population assumptions. These are scope limits, not additional validated domains.

When the deployment claim concerns unseen people, new-subject inference should be primary. Fixed-record inference is a conditional precision reference, not a mathematical lower bound for other targets. Intervals crossing zero do not establish equivalence.

Report the fields in Table~\ref{tab:2}, including eligibility counts and sensitivity analyses. Pairing, target weights, and the complete confusion matrix must be preserved inside resampling.

The audit is diagnostic rather than prescriptive: overlap may improve latency, temporal coverage, or point prediction, but denser segmentation does not manufacture an equal amount of independent evaluation evidence, and reporting nominal and information growth together makes that distinction visible.

\section{Limitations}
Validation covers two wearable HAR datasets and two classifier families, so the numerical factors are not universal; broader domains and models remain future work. The audit combines established estimators. HAC is bandwidth-dependent and conditional on observed sessions; it must not be relabeled as new-subject inference. Strong overlap and uneven recording lengths can leave residual anti-conservatism. HARTH's 22 clusters also limit the reliability of sandwich and percentile-bootstrap intervals.

Frozen OOF predictions are the evaluation object; we do not fully propagate training randomness across overlapping folds \cite{r11}, \cite{r12}. Within-session stationarity remains an approximation, and abrupt regime changes motivate segmentation or block-based sensitivity checks.

\section{Conclusion}
Sliding-window evaluation can transform a fixed amount of raw sensor data into many more test rows without a proportional increase in independent evidence. Reliable reporting therefore requires more than a subject-disjoint split: the target population, aggregation rule, highest independent unit, and metric-specific uncertainty method must be stated. Controlled and real-data results show that an explicit audit materially changes reported precision and can expose target- and metric-sensitive conclusions. The audit makes these checks reproducible from paired prediction files.

\textbf{Availability.} Code, frozen predictions, configurations, and reproduction scripts: \url{https://github.com/xinze8806-ship-it/sliding-window-confidence-audit}, release v1.0.0.

\bibliographystyle{IEEEtran}
\bibliography{references}

@article{r1,
  author = {O. Banos and J.-M. Galvez and M. Damas and H. Pomares and I. Rojas},
  title = {{Window size impact in human activity recognition}},
  journal = {Sensors}, volume = {14}, number = {4}, pages = {6474--6499}, year = {2014}
}

@article{r2,
  author = {A. Dehghani and T. Glatard and E. Shihab},
  title = {{Subject cross validation in human activity recognition}},
  journal = {arXiv preprint arXiv:1904.02666}, year = {2019}
}

@article{r3,
  author = {A. Dehghani and O. Sarbishei and T. Glatard and E. Shihab},
  title = {{A quantitative comparison of overlapping and non-overlapping sliding windows for human activity recognition using inertial sensors}},
  journal = {Sensors}, volume = {19}, number = {22}, pages = {5026}, year = {2019}
}

@article{r4,
  author = {A. Jord{\~a}o and Nazare, Jr., A. C. and J. Sena and W. R. Schwartz},
  title = {{Human activity recognition based on wearable sensor data: A standardization of the state-of-the-art}},
  journal = {arXiv preprint arXiv:1806.05226}, year = {2018}
}

@article{r5,
  author = {H. Bragan{\c c}a and J. G. Colonna and H. A. B. F. Oliveira and E. Souto},
  title = {{How validation methodology influences human activity recognition mobile systems}},
  journal = {Sensors}, volume = {22}, number = {6}, pages = {2360}, year = {2022}
}

@article{r6,
  author = {G. Coulaud and R. Akbarinia and F. Masseglia},
  title = {{Disjoint or overlapping? inference windowing for reconstruction-based time series anomaly detection}},
  journal = {arXiv preprint arXiv:2606.09874}, year = {2026}
}

@article{r7,
  author = {O. Binette and J. P. Reiter},
  title = {{Improving the validity and practical usefulness of AI/ML evaluations using an estimands framework}},
  journal = {arXiv preprint arXiv:2406.10366}, year = {2024}
}

@article{r8,
  author = {T. Hong and D. Lim and W. Bae},
  title = {{Beyond point estimates: Reliable evaluation of prediction performance metrics under clustered data}},
  journal = {arXiv preprint arXiv:2606.03656}, year = {2026}
}

@article{r9,
  author = {K. Anglin},
  title = {{Estimating uncertainty in classifier performance with applications to large language models and nested data}},
  journal = {arXiv preprint arXiv:2606.26422}, year = {2026}
}

@article{r10,
  author = {T. G. Dietterich},
  title = {{Approximate statistical tests for comparing supervised classification learning algorithms}},
  journal = {Neural Computation}, volume = {10}, number = {7}, pages = {1895--1923}, year = {1998}
}

@article{r11,
  author = {C. Nadeau and Y. Bengio},
  title = {{Inference for the generalization error}},
  journal = {Machine Learning}, volume = {52}, number = {3}, pages = {239--281}, year = {2003}
}

@article{r12,
  author = {Y. Bengio and Y. Grandvalet},
  title = {{No unbiased estimator of the variance of K-Fold cross-validation}},
  journal = {Journal of Machine Learning Research}, volume = {5}, pages = {1089--1105}, year = {2004}
}

@article{r13,
  author = {W. K. Newey and K. D. West},
  title = {{A simple, positive semi-definite, heteroskedasticity and autocorrelation consistent covariance matrix}},
  journal = {Econometrica}, volume = {55}, number = {3}, pages = {703--708}, year = {1987}
}

@article{r14,
  author = {H. R. K{\"u}nsch},
  title = {{The jackknife and the bootstrap for general stationary observations}},
  journal = {The Annals of Statistics}, volume = {17}, number = {3}, pages = {1217--1241}, year = {1989}
}

@article{r15,
  author = {D. N. Politis and J. P. Romano},
  title = {{The stationary bootstrap}},
  journal = {Journal of the American Statistical Association}, volume = {89}, number = {428}, pages = {1303--1313}, year = {1994}
}

@book{r16,
  author = {S. N. Lahiri}, title = {{Resampling Methods for Dependent Data}},
  address = {New York, NY, USA}, publisher = {Springer}, year = {2003}
}

@article{r17,
  author = {K.-Y. Liang and S. L. Zeger},
  title = {{Longitudinal data analysis using generalized linear models}},
  journal = {Biometrika}, volume = {73}, number = {1}, pages = {13--22}, year = {1986}
}

@article{r18,
  author = {A. C. Cameron and D. L. Miller},
  title = {{A practitioner's guide to cluster-robust inference}},
  journal = {Journal of Human Resources}, volume = {50}, number = {2}, pages = {317--372}, year = {2015}
}

@article{r19,
  author = {V. Saravanan and G. J. Berman and S. J. Sober},
  title = {{Application of the hierarchical bootstrap to multi-level data in neuroscience}},
  journal = {Neurons, Behavior, Data Analysis, and Theory}, volume = {3}, number = {5}, pages = {1--25}, year = {2020}
}

@article{r20,
  author = {T. P. Morris and I. R. White and M. J. Crowther},
  title = {{Using simulation studies to evaluate statistical methods}},
  journal = {Statistics in Medicine}, volume = {38}, number = {11}, pages = {2074--2102}, year = {2019}
}

@misc{r21,
  author = {G. M. Weiss},
  title = {{WISDM smartphone and smartwatch activity and biometrics dataset}},
  howpublished = {UCI Machine Learning Repository}, year = {2019}
}

@article{r22,
  author = {A. Logacjov and K. Bach and A. Kongsvold and H. B. B{\aa}rdstu and P. J. Mork},
  title = {{HARTH: A human activity recognition dataset for machine learning}},
  journal = {Sensors}, volume = {21}, number = {23}, pages = {7853}, year = {2021}
}

@inproceedings{r23,
  author = {A. Dempster and D. F. Schmidt and G. I. Webb},
  title = {{MiniRocket: A very fast (almost) deterministic transform for time series classification}},
  booktitle = {Proceedings of the 27th ACM SIGKDD Conference on Knowledge Discovery \& Data Mining},
  year = {2021}, pages = {248--257}
}
\end{document}